\documentclass[sigconf,nonacm]{acmart}
\AtBeginDocument{%
  }

\usepackage{multirow}
\usepackage{fontawesome}
\usepackage{caption}
\usepackage{subcaption}
\usepackage{booktabs}

\setcopyright{none}
\begin{document}

\title{HelixPipe: Efficient Distributed Training of Long Sequence Transformers with Attention Parallel Pipeline Parallelism}


\author{
{\rm Geng Zhang, Shenggan Cheng, Xuanlei Zhao, Ziming Liu, Yang You$^*$}
}
\thanks{$^*$Corresponding author: Yang You (youy@comp.nus.edu.sg)}
\affiliation{%
  \institution{National University of Singapore}
  \city{Singapore}
  \country{Singapore}
}

\renewcommand{\shortauthors}{Trovato et al.}

\begin{abstract}
As transformer sequence lengths grow, existing pipeline parallelisms incur suboptimal performance due to the quadratic attention computation and the substantial memory overhead.
To relieve these challenges, we propose HelixPipe, a novel pipeline parallelism for long sequence transformer training. 
First, HelixPipe introduces attention parallel partition, which schedules attention computations of different micro batches across different pipeline stages in parallel, reducing pipeline bubbles. 
Second, it employs a two-fold first-in-last-out micro batch schedule to balance memory usage and overlap communication with computation. 
Additionally, HelixPipe utilizes recomputation without attention and chunked MLP to mitigate fragmentation and enable longer sequences.
Experiments demonstrate that HelixPipe gains increasing advantages with longer sequence lengths, and outperforms existing methods in throughput and scalability across varying pipeline sizes, model sizes, and cluster configurations. 
Notably, it achieves a 26\% speedup over baseline methods when training a 7B model with 128k sequence length on 64 H20 GPUs.
Code is available at \url{https://github.com/code-tunnel/Megatron-LM/tree/dev}.
\end{abstract}

\maketitle

\section{Introduction}

Transformer models have rapidly advanced in handling longer sequence lengths, enhancing their understanding and generation capabilities in applications such as coding assistant \cite{guo2024deepseek}, image generation \cite{kong2024hunyuanvideo}, AI for science \cite{abramson2024accurate}. 
However, as sequence lengths grow, the quadratic complexity of attention computation and the substantial memory footprint pose significant challenges for distributed training of transformer models.
Current approaches often employ multiple orthogonal parallelisms, including data parallelism \cite{xing2015petuum}, tensor parallelism \cite{shoeybi2019megatron}, sequence parallelism \cite{jacobs2023deepspeed} and pipeline parallelism \cite{narayanan2021memory}, to train transformer models efficiently \cite{zheng2022alpa, yuan2024accelerating}.
While a variety of sequence parallelisms, such as Megatron sequence parallelism \cite{korthikanti2023reducing}, Deepspeed Ulysses \cite{jacobs2023deepspeed}, Ring attention \cite{liu2023ring, li2023sequence}, Megatron context parallelism \cite{context_parallel}, USP \cite{fang2024unified} have been proposed to split individual activations along the sequence dimension across GPUs to alleviate the computation and memory overhead, the impact of long sequence lengths on the computation efficiency and memory footprint for pipeline parallelisms remains underexplored \cite{lin2025weipipe, kim2023bpipe}.

Existing pipeline parallelisms, such as 1F1B \cite{narayanan2021memory, fan2021dapple, llama3} and zero bubble pipeline parallelism \cite{qi2024zero, deepseekv2}, were designed with sequence lengths below 2048.
These methods partition a transformer model to a set of consecutive layers, and maps each set to a single pipeline stage.
For each training iteration, the input data batch is split to multiple micro batches to execute the model along the pipeline stages.
While 1F1B executes one forward pass and one backward pass at a time for each stage, zero bubble pipeline parallelism decouples the backward pass, and executes one forward pass, one backward for input activations and one backward for model parameters to reduce the pipeline bubble and improve the computation efficiency.

Though existing pipeline parallelisms perform well with short sequence lengths, they achieve suboptimal performance with longer sequence lengths for two reasons.
First, because of the data dependency between pipeline stages and the optimizer synchronization, pipeline parallelisms incur inevitable pipeline bubbles to downgrade its computation efficiency.
Since existing methods partition a transformer model by layers, the pipeline bubble is proportional to the execution time of transformer layers.
Due to the quadratic complexity of attention computation, pipeline bubble will be dominated by the attention computation as the sequence lengths increase, leaving substantial room for improvement.
Second, following the 1F1B micro batch schedule, each pipeline stage needs to stash the activations for a different number of micro batches before they are used and released in the later backward pass.
Such an imbalanced memory footprint is detrimental for training long sequence transformer models, as later pipeline stages can have low memory utilization whereas the earlier pipeline stages run out of memory.
While zero bubble pipeline parallelism improves the computation efficiency, the decoupled backward pass leads to the same peak memory as 1F1B for all pipeline stages.

To address these challenges, we present HelixPipe, a novel pipeline parallelism for efficient transformer training with long sequence lengths.
Given the observation that attention computation of a transformer layer does not rely on model parameters, HelixPipe proposes the \textit{attention parallel partition}.
It schedules the attention computation of the same layer for different micro batches to different pipeline stages, so that the attention computations are executed in parallel across pipeline stages and can be removed from the pipeline bubbles to improve the computation efficiency.

To balance the memory footprint across pipeline stages, HelixPipe proposes a \textit{two-fold first-in-last-out (FILO) micro batch schedule}.
The FILO schedule first executes the forward pass for all micro batches and then finishes their backward pass in the reverse order, so that it can stash the activations for the same number of micro batches at each stage before executing their backward pass to balance the peak memory usage.
Moreover, as there is not data dependency between micro batches, instead of executing one micro batch at a time, HelixPipe executes two micro batches (fold) every time to hide the communication of one micro batch with the computation of the another.

In addition, HelixPipe further optimizes the memory usage by a recomputation without attention strategy along with chunked MLP.
As the attention computation dominates the execution time per layer, HelixPipe recomputes the intermediate activations of all operators except for the attention before the backward pass, significantly reducing the memory overhead at a marginal performance drop with the increase of sequence length.
HelixPipe adopts chunked MLP to mitigate memory fragmentation introduced by the joint impact from the recomputation strategy, the two-fold FILO schedule and the long sequence lengths.

The contributions of this paper are summarized as follows:
\begin{itemize}
    \item Attention parallel partition is proposed to execute the attention computation of the same layer for different micro batches in parallel across pipeline stage to remove the heavy attention computation from pipeline bubble.
    \item Two-fold FILO micro batch schedule is proposed to balance the memory footprint across pipeline stages and overlap the communication with computation.
    \item Recomputation without attention strategy with chunked MLP is designed to optimize the practical memory utilization to extend the sequence length.
    \item Empirical results show that HelixPipe has higher throughput and better scalability than baseline methods across various sequence lengths, pipeline size, model scales and GPU types. In particular, it exceeds the best baseline by 26\% for training a 7B model with 128k sequence length and 64 H20 GPUs.
\end{itemize}

\section{Preliminaries}
\subsection{Model Architecture}
\begin{figure}[!t]
    \centering
    \includegraphics[width=\columnwidth]{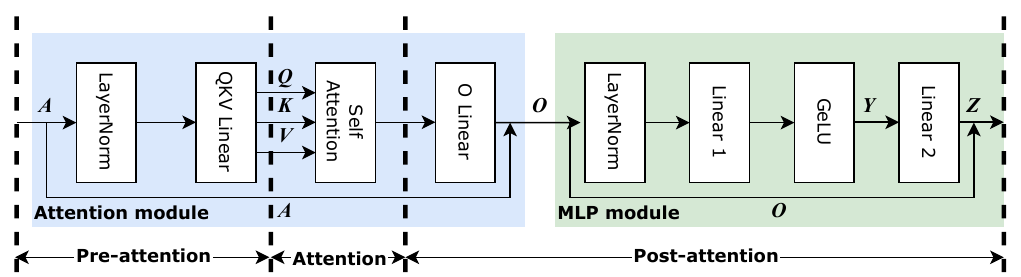}
    \caption{A transformer layer. The layer can be partitioned into three parts: pre-attention, attention, post-attention. Only the attention part is non-parameterized.}
    \label{fig:transformer-block}
\end{figure}

A transformer model mainly comprises $L$ transformer layers.
This section and the following of this paper take the standard model architecture of GPT-3 \cite{brown2020language} as the targeting model for analysis.
The details of a transformer layer are depicted in Figure \ref{fig:transformer-block}.
Given the input activation $A$ of shape $[s, b, h]$, where $s$ is the sequence length, $b$ is the micro batch size, $h$ is the hidden size of the model, the computation and memory overhead of each operation are summarized in Table \ref{transformer-block} following convention \cite{narayanan2021efficient, korthikanti2023reducing}.

\subsection{Sequence Parallelism}
Sequence parallelisms split individual tensors across multiple devices to reduce the memory overhead at the \textit{intra-layer level} \cite{zheng2022alpa}.
There are a number of sequence parallelisms, such as Megatron sequence parallelism \cite{korthikanti2023reducing}, Deepspeed Ulysses \cite{jacobs2023deepspeed}, Ring attention \cite{liu2023ring, li2023sequence}, Megatron context parallelism \cite{context_parallel}, USP \cite{fang2024unified}, etc.
This work and the baseline methods utilizes Megatron-LM sequence parallelism for scaling the sequence length at the intra-layer level, as it can partition individual model parameters and activations\cite{korthikanti2023reducing}.

For the attention module, the input activation is first split along $s$ dimension across $t$ GPUs.
Then, an all-gather operation is used to recover the full sequence to perform partial attention computation with partitioned model parameters.
After the local computation, the activation is reduced and split via a reduce-scatter operation.
The MLP module has the same communication pattern for the input and output activation to enable the sequence parallelism.
In the backward pass, the two communication operations are reversed for the gradients.
As the sequence parallelism relies on frequent collective communication for each transformer layer, it is often applied within a GPU server to exploit the high bandwidth of NVLink.

\begin{table*}[!t]
\caption{The computation and memory overhead of a transformer layer.
The FLOPs of matrix computations is reported as the computation overhead.
Backward \textit{B} and \textit{W} denotes the gradient computation for input activations and model parameters respectively.
The number of elements in the model states and activations are separately reported as the memory overhead. Bias parameters are neglected for brevity.
Intermediate data of attention is rounded to $3bsh$ due to flash attention \cite{dao2022flashattention}. 
Dropout is omitted due to low memory dropout \cite{lowmemdropout}.
}
\label{transformer-block}
{\small
\begin{tabular*}{\textwidth}{@{\extracolsep{\fill}}l|cccc|cccc|l}
\hline
             & \multicolumn{4}{c|}{Attention module}                                                                                          & \multicolumn{4}{c|}{MLP module}                                                                                         & \multicolumn{1}{c}{\multirow{2}{*}{Total}} \\ \cline{2-9}
             & \multicolumn{1}{c}{LayerNorm} & \multicolumn{1}{c}{QKV Linear} & \multicolumn{1}{c}{Attention} & \multicolumn{1}{c|}{O Linear} & \multicolumn{1}{c}{LayerNorm} & \multicolumn{1}{c}{Linear 1} & \multicolumn{1}{c}{GeLU} & \multicolumn{1}{c|}{Linear 2} & \multicolumn{1}{c}{}                       \\ \hline
Forward  & -                             & $6bsh^2$                       & $4bhs^2$                      & $2bsh^2$                      & -                             & $8bsh^2$                     & -                        & $8bsh^2$                      & $4bsh(6h+s)$                               \\
Backward \textit{B} & -                           & $6bsh^2$                       & $8bhs^2$                      & $2bsh^2$
& -                             & $8bsh^2$                     & -                        & $8bsh^2$                      & $4bsh(6h+2s)$
                            \\
Backward \textit{W} & -                           & $6bsh^2$                       & -                      & $2bsh^2$
& -                             & $8bsh^2$                     & -                        & $8bsh^2$                      & $4bsh(6h)$
                            \\
Model parameters & $2h$                          & $3h^2$                      & -                             & $h^2$                       & $2h$                          & $4h^2$                    & -                        & $4h^2$                      & $12h^2+4h$                                \\
Activation   & $bsh$                           & $bsh$                            & $3bsh$                          & $bsh$                           & $bsh$                           & $bsh$                          & $4bsh$                     & $4bsh$                          & $16bsh$                                    \\ \hline
\end{tabular*}
}
\end{table*}

\begin{figure*}[!t]
    \centering
    \subfloat[1F1B pipeline schedule.]{\label{fig:1f1b-4b}
    \centering
    \includegraphics[width=\linewidth]{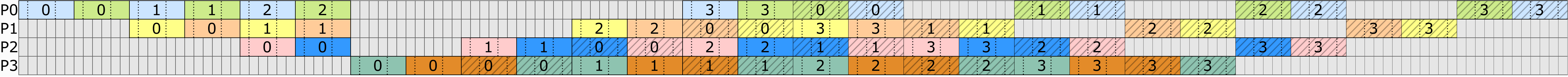}
    }
    \hfill
    \subfloat[HelixPipe FILO schedule.]{\label{fig:helix-4b}
    \centering
    \includegraphics[width=\linewidth]{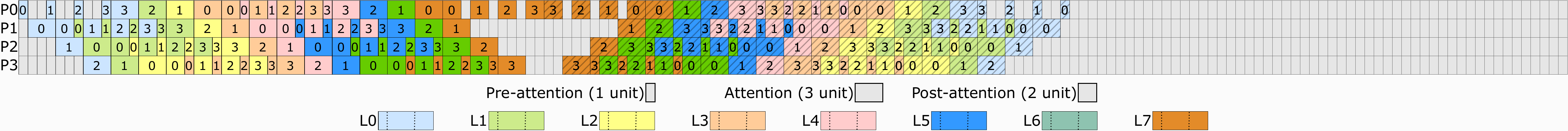}
    }
    
    \caption{1F1B and HelixPipe FILO schedules for 4 micro batches (numbered) execute 8 layers over 4 pipeline stages. Backward execution time in shadow is shown equivalent to forward for brevity. Execution time ratio of pre-attention, attention, and post-attention is set to 1:3:2.}
    \label{fig:pipe-design}
\end{figure*}

\subsection{Pipeline Parallelism} \label{pp-background}
Pipeline parallelisms scale the distributed training of transformer models at the \textit{inter-layer level}.
Various pipeline parallelisms have been proposed \cite{narayanan2021efficient, kim2023bpipe, sun2024adapipe, lin2025weipipe, huang2019gpipe}.
This section takes 1F1B \cite{narayanan2021memory, fan2021dapple} and zero bubble pipeline parallelism \cite{qi2024zero} for discussion, as 1F1B is the most commonly adopted pipeline parallelism \cite{llama3, bi2024deepseek} and zero bubble pipeline parallelism demonstrates high efficiency for training large scale transformer models \cite{deepseekv2, liu2024deepseek}.
In particular, this work takes ZB1P, one of the variants of zero bubble pipeline parallelism for discussion, as it costs the same peak memory as 1F1B\footnote{Another variant of zero bubble pipeline  parallelism is ZB2P, which costs more memory and involves optimizer modification.}.
This section briefly reviews the two pipelines and quantifies their pipeline bubble and memory overhead.
The discussion of other pipeline parallelisms is extended to Section \ref{related-work}.

\subsubsection{1F1B Pipeline}
Given an $L$-layer transformer model, 1F1B evenly partitions the model and assigns to $p$ pipeline stages, each with $L/p$ consecutive layers.
In a training iteration, a data batch of $B$ samples is split into $m$ micro batches to execute the pipeline.
The micro batch schedule is visualized in Figure \ref{fig:1f1b-4b}.
It first executes forward passes of $p-i-1$ micro batches to warm up the $i$-th pipeline stage.
Then it enters the steady phase to execute a forward pass and a backward pass (so called 1F1B) each time for the remaining micro batches.
Finally, the backward passes of all outstanding micro batches are finished.
Due to the lightweight need for transporting activation and gradients via p2p communication between pipeline stages, each pipeline stage is often mapped to a GPU server, and the p2p relies on fast inter-node channels like Infiniband \cite{narayanan2021efficient}.

\textbf{Pipeline bubble.}
As visualized in Figure \ref{fig:1f1b-4b}, the pipeline bubble equals to $p-1$ forward and backward executions of $L/p$ transformer layers in each stage.
Let $t_{pre} + t_{attn} + t_{post}$ denote the layer forward time, in which the three items denote the time of pre-attention, attention and post-attention respectively.
According to Table \ref{transformer-block}, the backward pass including the gradient computations for input activations (backward \textit{B}) and model parameters (backward \textit{W}) requires approximately twice the computational cost of the forward pass for each layer.
Thus, the bubble time in an iteration can be quantified as:
\begin{align} \label{comp-1f1b}
    T^{1F1B} =3(p-1)(t_{pre} + t_{attn} + t_{post})L/p
\end{align}

\textbf{Activation memory.}
For pipeline stage $i$, the number of outstanding micro batches that reserve activations reaches peak at the end of the first forward pass of the steady phase, that is, $p-i$.
Thus, the activation memory overhead at pipeline stage $i$ can be formulated as:
\begin{align} \label{mem-1f1b}
    M_{i}^{1F1B}=16(p-i)bshL/p
\end{align}
Note that for the first stage where $i=0$, the activation overhead is $16bshL$, which is irrelevant to pipeline size $p$.

\subsubsection{ZB1P Pipeline}
ZB1P is built upon 1F1B.
Based on the observation that backward pass only relies on the backward \textit{B} to fulfill the data dependency, ZB1P decouples backward pass into backward \textit{B} and backward \textit{W} , and delays the backward \textit{W} to fill the pipeline bubble at the final phase when finishing outstanding micro batches.
Therefore, ZB1P inherits the layer-wise model partition of 1F1B, and the micro batch schedule follows a one forward, one backward \textit{B} and one backward \textit{W} pattern.

Moreover, ZB1P can achieve its best performance when the forward, backward \textit{W} and backward \textit{B} cost the same amount of time, which may be false in real practice.
To adapt to various computation time, ZB1P combines a heuristic algorithm with integer linear programming to automatically search for the optimal backward \textit{W} schedule to fill the pipeline bubble at the same memory capacity as 1F1B for all pipeline stages.

\textbf{Pipeline bubble.}
ZB1P reduces pipeline bubble by delaying the backward \textit{W} to fill the bubble of 1F1B.
Formally, its pipeline bubble is quantified as:
\begin{align} \label{comp-zb1p}
    T^{ZB1P} =(p-1)(t_{pre} + 3t_{attn} + t_{post}) L/p
\end{align}
Note that because the attention computation does not involve model parameters and is not included in backward \textit{W}, backward \textit{W} can only reduce pipeline bubble by the pre-attention and post-attention computation.

\textbf{Activation memory.}
Due to the need of delaying backward \textit{W}, the intermediate activations must be saved until the execution of backward \textit{W} for each micro batch.
Thus, the memory overhead of ZB1P at the worst case for each pipeline stage is the same as the peak memory overhead of 1F1B:
\begin{align} \label{mem-zb1p}
    M^{ZB1P}=16bshL
\end{align}

\section{Motivation}
While existing pipeline parallelisms work well with existing transformer models whose sequence length are typically set to 2k or 4k \cite{narayanan2021efficient, qi2024zero},
they incur suboptimal computation efficiency and substantial memory footprint with the increase of sequence length.

\subsection{Attention Dominated Pipeline Bubble} \label{motivation:compute}
\begin{figure}[!t]
    \centering
    \includegraphics[width=0.78\columnwidth]{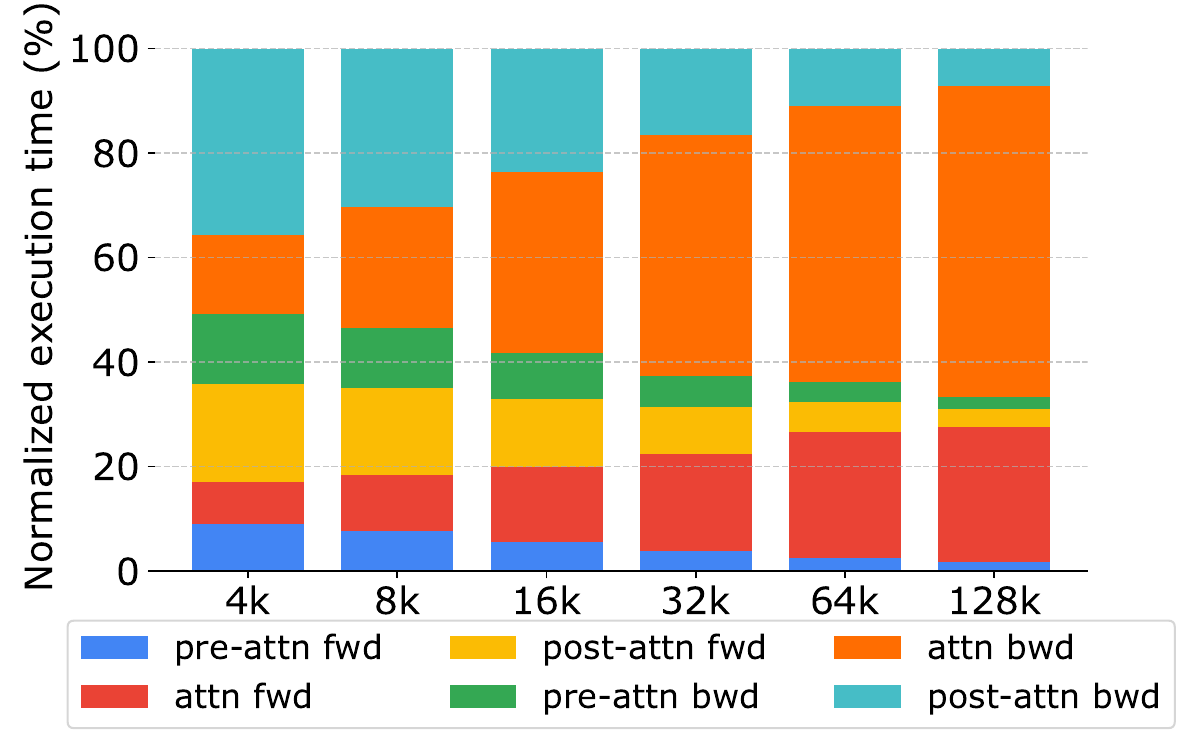}
    \caption{The normalized duration for each component of a transformer layer with various sequence lengths profiled on an A800 GPU. $h$ is 4096. $b$ is 1. Flash attention is enabled \cite{dao2022flashattention}.}
    \label{fig:block-profile}
\end{figure}

Existing pipeline parallelisms such as 1F1B or ZB1P partition a model at the layer granularity.
This makes the pipeline bubble of 1F1B proportional to the execution time of a transformer layer as shown in Equation \ref{comp-1f1b}.
For ZB1P, since the backward \textit{W} excludes the non-parameterized attention module, the decoupled backward pass does not affect attention computation in the bubble in Equation \ref{comp-zb1p}.
However, \textbf{\textit{with long sequences, the attention computation with quadratic complexity dominates execution time}}, as shown by the profiling results in Figure \ref{fig:block-profile}.
It breaks down forward and backward time into pre-attention, attention, and post-attention phases. 
Since the bubble scales with layer execution time, attention becomes the primary bottleneck in long sequence scenarios. Addressing this issue could significantly improve pipeline computation efficiency.

Theoretically, the pipeline bubble overhead can be minimized to near zero with larger global batch size and more micro batches.
However, processing an extremely large number of tokens per iteration can hurt the convergence and lead to longer training time \cite{You2020Large}.
Instead, the global number of tokens per each training iteration is fixed in practice.
For instance, both Llama 1 and Llama 2 maintain 4M tokens per iteration \cite{touvron2023llama, touvron2023llama2}, while Llama 3 405B is trained with 16M tokens for each iteration \cite{llama3}.
\textbf{\textit{When working with long sequence lengths, this fixed token budget constrains the batch size and the number of micro batches available for pipeline parallelism}}.
For example, Llama 3 405B with 128k sequence length is trained using a batch size of 16 and 16 pipeline stages.
Such a pipeline is merely saturated, further amplifying the inefficiency introduced by the pipeline bubble.

\subsection{Memory Imbalance across Pipeline Stages} \label{mot:balance}
Pipeline parallelism can evenly reduce the memory pressure from model states by splitting transformer layers into multiple pipeline stages.
However, as quantified by Equation \ref{mem-1f1b}, 1F1B cannot reduce the activation memory footprint and introduces a memory overhead imbalance across different pipeline stages.
According to Equation \ref{mem-zb1p}, ZB1P maintains the same worst-case peak memory usage as 1F1B for all pipeline stages.
With long sequence length, the skewed activation memory overhead becomes the bottleneck for training transformer models.

\begin{figure}[!t]
    \centering
    \includegraphics[width=0.8\columnwidth]{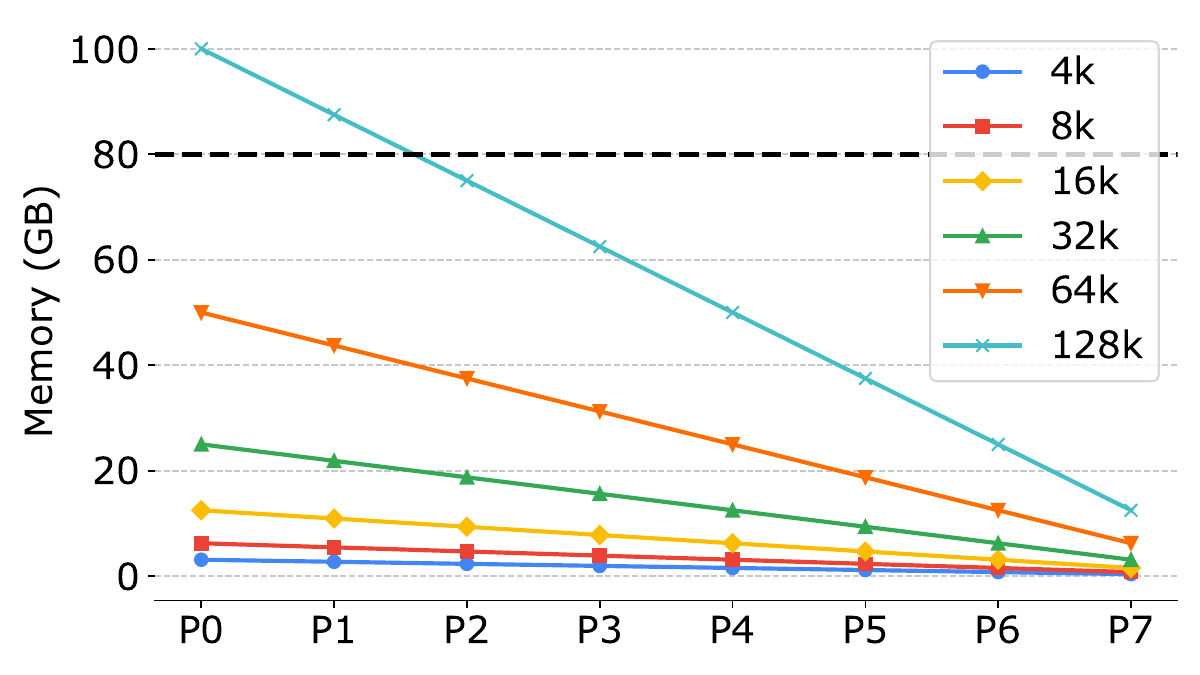}
    \caption{Activation memory overhead for a 13B transformer with 8 pipeline stages at different sequence length. 1F1B schedule and FP16 are used.}
    \label{fig:actmem}
\end{figure}

Figure \ref{fig:actmem} shows the theoretical activation memory overhead for 1F1B training a transformer model with 13B parameters and eight GPU nodes, each with eight A800 GPUs.
According to this figure, when sequence length increases to 128k, the activation memory demands at the first and the second stages exceed the 80G GPU memory capacity.
However, later pipeline stages leave large spare memory.
Though shorter sequence lengths will not exceed the memory capacity, they may not run in practice because of other memory required by model states and intermediate data \cite{rajbhandari2020zero, yuan2024accelerating}.
This implies that \textbf{\textit{while increasing pipeline size can reduce the model states in each stage, it cannot relieve the memory pressure costed by activation.}}
Therefore, memory usage should be carefully considered to design an efficient pipeline parallelism under longer sequence length.
While a few methods, such as AdaPipe \cite{sun2024adapipe}, BPipe \cite{kim2023bpipe} have been proposed for this memory issue, they do not take the pipeline bubble time into consideration.

\section{HelixPipe} \label{attn-pipe-schdl}
\subsection{Overview}
This work proposes HelixPipe to alleviate the pipeline bubble and memory imbalance for long sequence transformer model training.
HelixPipe adopts a fine grained \textit{attention parallel partition} to remove the heavy attention computation from pipeline bubble.
To efficiently harness GPU memory across pipeline stages and relieve the communication overhead, HelixPipe proposes a \textit{two-fold first-in-last-out micro batch schedule}.
Beyond the two core designs, HelixPipe utilizes activation recomputation without attention along with chunked MLP to further optimize the memory utilization.

We note that while HelixPipe schedules the execution of different micro batches for different layer components, it preserves the computation order for individual micro batches in each iteration.
Thus it maintains the same computation semantics and convergence as 1F1B or ZB1P.

\subsection{Attention Parallel Partition} \label{attn-parallel-partition}
In conventional pipeline parallelisms such as 1F1B and ZB1P, transformer models are partitioned at the layer granularity in each stage, making the attention computation of a sequence of micro batches also sequential, and the pipeline bubble proportional to the execution time of a whole layer.
Exploiting the observation that attention computation is not parameterized, this work introduces the attention parallel partition, which schedules the attention computation of multiple micro batches across the pipeline stages.
By executing the attention computation of different micro batches in parallel, attention parallel partition can remove the attention from pipeline bubble and improve the efficiency of pipeline.

Specifically, each transformer layer in HelixPipe is partitioned to three parts, pre-attetion, attention and post-attention as shown in Figure \ref{fig:transformer-block}.
Since only pre-attention and post-attention have parameters, the attention parallel partition maps them to pipeline stages in a helix pattern (so called HelixPipe).
First, the pre-attention of $0$-th layer is assigned to stage $0$.
Then, for $l\in [1, L)$, post-attention of layer $(l-1)$ and pre-attention of layer $l$ are concatenated to stage $(l \bmod p)$.
Finally, the post-attention of the last layer $L-1$ is mapped into stage $0$.
For attention part, the micro batch $i$ of layer $l$ is assigned to stage $((l+i+1) \bmod p)$, which makes different attention computation executed in parallel.

Between the boundary of attention and post-attention, the activations or gradients for attention output and residual input are transferred with the communication volume of $2bsh$.
However, for the boundary of pre-attention and attention, attention input, $\textbf{Q}$, $\textbf{K}$ and $\textbf{V}$ and residual input $\textbf{A}$ are communicated. 
The overall communication volume of this boundary is $4bsh$.
To relieve the heavy data volume between the pre-attention and attention, instead of transferring $\textbf{Q}$, $\textbf{K}$ and $\textbf{V}$, HelixPipe transfers the QKV linear parameters along with its input $\textbf{A}$, moving its computation to attention part.
This reduces the data volume to $2bsh+3h^2$.
For a long sequence length where $s \gg h$, the data volume is approximately $2bsh$.

\begin{figure}[!t]    
    \centering
    \subfloat[Layer-wise partition.]{\label{fig:block-par}
    \centering
    \includegraphics[width=0.45\columnwidth]{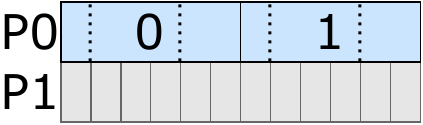}
    }
    \hfill
    \subfloat[Attention parallel partition.]{\label{fig:helix-par}
    \centering
    \includegraphics[width=0.45\columnwidth]{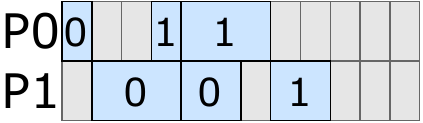}
    }
    \caption{Layer-wise model partition in existing pipeline and attention parallel partition in HelixPipe for one layer in a two-stage pipeline and two micro batches (numbered). Execution time ratio of pre-attention, attention, and post-attention is set to 1:3:2.}
    \label{fig:partition}
\end{figure}

An example for the attention parallel partition of HelixPipe, and its difference from layer-wise model partition is depicted in Figure \ref{fig:partition}.
In this example and the following, the ratio of execution time for pre-attention, attention, and post-attention is set to 1:3:2 for discussion.
A transformer layer is mapped to a two-stage pipeline, and there are two micro batches to saturate the pipeline.
Figure \ref{fig:partition} indicates that HelixPipe is more efficient than existing pipeline parallelisms.
The benefit comes from executing the attention of different micro batches in parallel and priorities the computation of following micro batches.
Longer sequences make the attention part larger, and are more advantageous for HelixPipe.

\subsection{Two-fold First-in-last-out (FILO) Schedule}
\subsubsection{FILO schedule}
To coordinate with the attention parallel partition and balance the activation memory overhead, HelixPipe adopts a first-in-last-out (FILO) schedule for executing micro batches.
The FILO schedule executes micro batches using multiple \textit{loops}\footnote{Here uses loop to differentiate from training iteration.}, with each loop admitting $p$ micro batches.
For each layer, these micro batches first executes the pre-attention part sequentially, and then are distributed to corresponding stages with p2p communication to execute attention in parallel.
In the following, the attention outputs are transferred to targeting pipeline stages and executed for the post-attention part before proceeding to next layer.
Since the maximum number of micro batches to be executed in parallel is equal to the number of pipeline size $p$, the FILO schedule requires that the number of micro batches $m$ must be divisible by $p$.
An example of the FILO scheduling four micro batches across four pipeline stages for an eight-layer transformer model is illustrated in Figure \ref{fig:helix-4b}.
This figure indicates that the FILO schedule has better computation efficiency than 1F1B.
However, this naive FILO schedule does not take communication into consideration.

\begin{figure}[!t]
    \centering
    \subfloat[Naive FILO schedule]{\label{fig:1-fold}
    \makebox[\linewidth][l]{%
    \includegraphics[height=70pt,keepaspectratio]{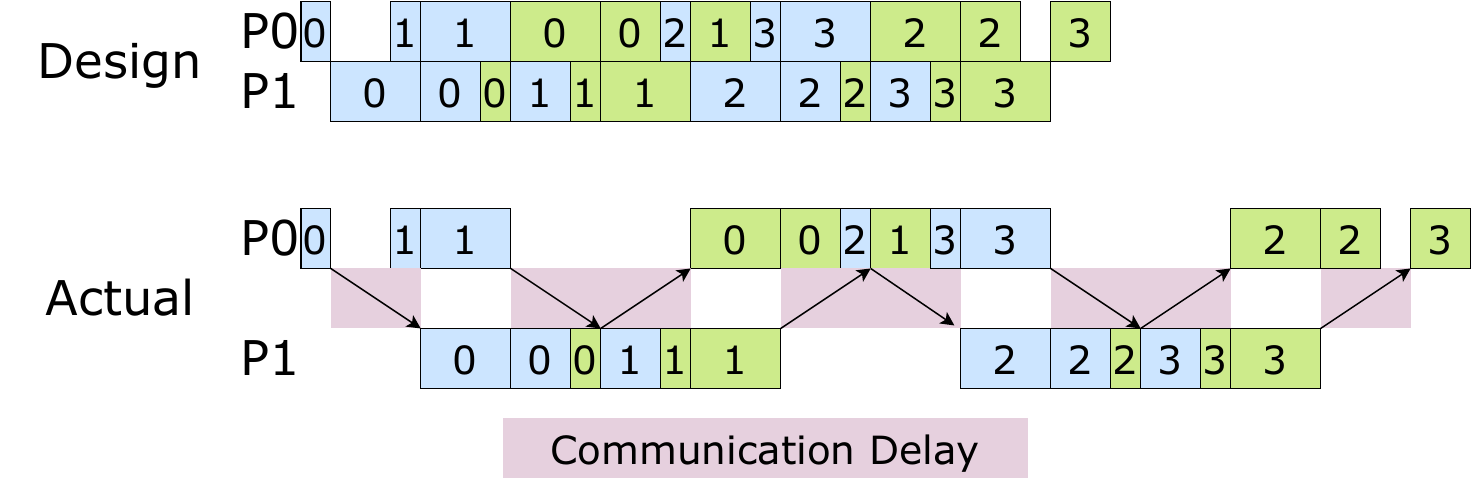}%
    }%
    }
    \hfill
    \subfloat[Two-fold FILO schedule]{\label{fig:2-fold}
    \makebox[\linewidth][l]{%
    \includegraphics[height=70pt,keepaspectratio]{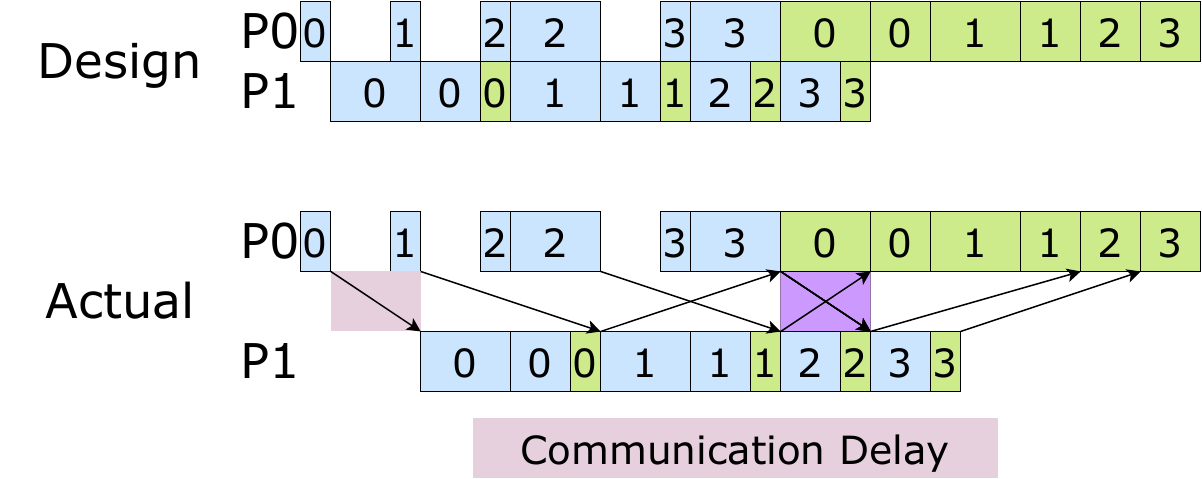}%
    }%
    }
    \caption{Two-fold FILO schedule overlaps the communication bottlenecks the naive FILO schedule.}
\end{figure}

\begin{figure*}[!t]
    \centering
    \subfloat[HelixPipe with the naive FILO schedule]{\label{fig:4-1-fold}
    \includegraphics[width=\linewidth]{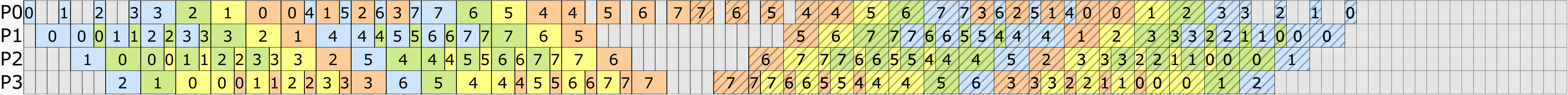}%
    }
    \hfill
    \subfloat[HelixPipe with the two-fold FILO schedule]{\label{fig:4-2-fold}
    \includegraphics[width=\linewidth]{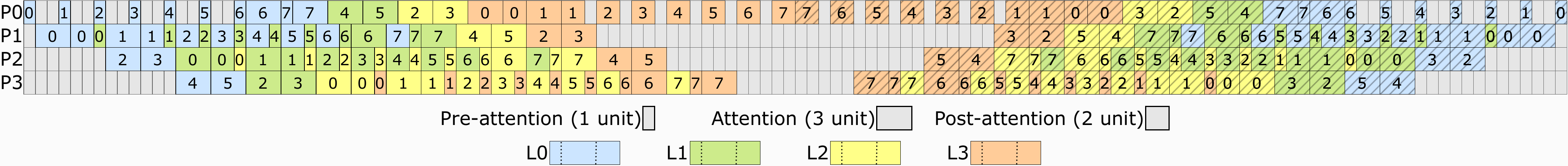}%
    }
    \caption{HelixPipe with naive and two-fold FILO schedules for 8 micro batches (numbered) execute 4 layers over 4 pipeline stages. Backward execution time in shadow is shown equivalent to forward for brevity.}
    \label{full-helix}
\end{figure*}

\subsubsection{Asynchronous Two-fold FILO Schedule}
For 1F1B or ZB1P, every $L/p$ layers only needs to transfer activations or gradients of size $bsh$. 
However, the attention parallel partition of HelixPipe demands a communication volume of $2bsh$ for each layer.
This frequent and heavy communication can cause nontrivial bottleneck along the critical path for the naive FILO schedule to decrease GPU utilization and computation efficiency.
Figure \ref{fig:1-fold} visualizes this problem with a pipeline of two stages.
The receiving at P1 delays its following sending, which in return delays the receiving at P0 and the following computation.

To resolve this bottleneck, the original naive FILO schedule can be enhanced to an asynchronous two-fold version.
The key observation is that, instead of executing one micro batch at a time, we can executing two micro batches (folds). 
Because there is no data dependency between the two micro batches, the two fold schedule offers asynchronous opportunities to overlap the communication.
A two-fold example is presented in Figure \ref{fig:2-fold} to clarify this idea.
While at the initial phase, there will be a communication delay, the following communication of one micro batch can be overlapped via the computation of another micro batch. 
As a result, the communication would not delay the pipeline computation.

Figure \ref{full-helix} compares the naive FILO schedule and the two-fold FILO schedule.
It indicates the two-fold FILO schedule addresses the communication bottleneck at the cost of double pipeline bubble time as it executes two micro batches at a time.
A quantitative analysis for the pipeline bubble time of HelixPipe is summarized in Section \ref{quant}.

The two-fold schedule also implies a demand for high communication bandwidth to ensure that the communication can be totally hidden by computation.
We will show in experiments in Section \ref{comm impact} that whether such communication can be overlapped is decided by if the attention computation can overlap the communication behind it, as depicted in the purple block of Figre \ref{fig:2-fold}, and existing hardware such as Infiniband can largely or even perfectly achieve this goal, allowing HelixPipe to scale to clusters of any size.

%

\subsection{Memory Optimization} \label{mem-opt}
\subsubsection{Recomputation without Attention} \label{recomp}
Motivated by the observation from Figure \ref{fig:block-profile} that the pre-attention and post-attention consume a marginal fraction of layer execution time with long sequence lengths, HelixPipe utilizes an activation recomputation without attention strategy to effectively reduce the memory overhead with a small cost of computation efficiency.

Specifically, during the forward pass, instead of stashing all the activations for backward pass, HelixPipe only stashes the input activations for the attention part and the combined post-attention and pre-attention, discarding all the intermediate activations.
Before the backward computation, HelixPipe uses the stashed activations to recompute the intermediate activations.
There are only two types of stashed activations for each transformer layer.
The first type is the input and output activations saved by flash attention for the attention computation \cite{dao2022flashattention}.
The input is $bsh+3h^2$ due to the optimization in Section \ref{attn-parallel-partition}, and the output is $bsh$.
To simplify notations, let the total amount be approximated to $2bsh$. 
The second type is the activations of size $2bsh$ for the combined pre-attention and post-attention, including the residual input and the attention output received from other stages.
Therefore, the activation memory required by each layer is reduced to $4bsh$.

It is noteworthy that the recomputation without attention strategy is similar to the one used in DistFlashAttn \cite{li2023lightseq}, and is totally opposite to the selective recomputation strategy \cite{korthikanti2023reducing}.
However, we identified that the strategy cannot be directly applied in practice due to memory fragmentation issues caused by it. 
Therefore, HelixPipe boosts the memory utilization with the following technique, chunked MLP to mitigate this issue.


\subsubsection{Chunked MLP} \label{chunk-mlp}
While recomputation without attention reduces memory usage, we observed severe memory fragmentation due to irregular allocations in MLP computations, worsened by long sequences and the two-fold FILO schedule. To mitigate this, we introduce chunked MLP, which processes MLP operations in smaller, manageable segments, and the caveat is to coordinate the two communications, all-gather and reduce-scatter operations for the sequence parallelism \cite{korthikanti2023reducing}.

Given sequence parallel size $t$, the MLP input activation $\textbf{O}$ is of shape $[s/t, b, h]$ at each GPU.
An all-gather recovers the full sequence $[s, h]$, but instead of processing it at once, we split it into chunks of size $[c, b, h]$, where $c$ is a configurable chunk size. This reduces peak memory fragmentation. After processing, the chunked outputs are concatenated into $[s, b, h/t]$, followed by a reduce-scatter to produce the final output $\textbf{Z}$ of shape $[s/t, b, h]$.
The memory usage is further optimized by pre-allocating reusable buffers for all-gather and reduce-scatter communications, eliminating dynamic memory overhead. 
The backward pass follows the same chunked strategy for gradient computation. 
This approach ensures stable memory usage while maintaining efficiency.

\subsection{Quantitative Analysis} \label{quant}
\begin{table}[!t]
    \centering
    \caption{Quantitative analysis for different pipeline parallelisms.}
\begin{tabular}{@{}lccccc@{}}
\toprule
Pipeline   & Pipeline bubble time                   & Activation memory \\ \midrule
1F1B       & $3(p-1)(t_{pre}+t_{attn}+t_{post})L/p$    & $16(p-i)bshL/p$     \\
ZB1P       & $(p-1)(t_{pre}+3t_{attn}+t_{post})L/p$    & $16bshL$            \\
HelixPipe  & $8(p-1)(t_{pre}+t_{post})$             & $4bshmL/p$          \\ \bottomrule
\end{tabular}
    \label{tab:quant_helix}
\end{table}

This section quantifies the pipeline bubble time and activation memory footprint for HelixPipe step by step.
The overall results of HelixPipe and the comparison to 1F1B or ZB1P are summarized in Table \ref{tab:quant_helix}.

According to Figure \ref{fig:4-1-fold}, with the naive FILO schedule, each pipeline stage incurs $p-1$ idle units for both the pre-attention and post-attention in both forward and backward pass.
As a result, for HelixPipe with the naive FILO schedule, the pipeline bubble time is equal to $3(p-1)(t_{pre}+t_{post})$, \textit{successfully removing the dominated attention computation from the pipeline bubble time and thus improving the computation efficiency compared to existing methods}.
Because each stage still has $L/p$ layers and each layer needs to stash $16bsh$, the activation memory footprint is $16bshmL/p$.

According to Figure \ref{fig:4-2-fold}, the two-fold FILO schedule executes two micro batches at a time to overlap the communication, so the pipeline bubble time for HelixPipe with the two-fold FILO schedule is doubled to $6(p-1)(t_{pre}+t_{post})$.
While the activation memory overhead does not change, the two-fold FILO schedule demands twice the number of micro batches to saturate the pipeline.

By introducing the recomputation without attention strategy, we need to rerun the forward pass of the pre-attention and post-attention units before their backward pass to recompute the required activations, so the pipeline bubble time now becomes $8(p-1)(t_{pre}+t_{post})$.
However, the activation memory can be reduced to $4bshmL/p$.
In other words, \textit{the recomputation strategy sacrifices $1/3$ pipeline bubble time but reduces the activation memory footprint by $4$ times}. 

\subsection{Implementation} \label{impl-opt}
HelixPipe is implemented with around 3,400 LoC, and is based on Megatron-LM to utilize data, tensor and sequence parallelisms to enable long sequence training. 
There are three major components, layer modules, pipeline schedule, pipeline context.
Layer modules rewrite the transformer layer according to the pre-attention, attention and post-attention partition, and implement the memory optimization discussed in Section \ref{mem-opt}.
Pipeline schedule implements the two-fold FILO schedule of HelixPipe.
Pipeline context manages the underlying communication and the memory buffers for sending and receiving micro batches.

In addition to the three components, there are two additional modules that need optimization.
In the end-to-end training of a large language model, there are two additional modules: (1) word \& position embeddings before the transformer layers and (2) word embedding for next token prediction and loss computation after the transformer layers.
For (1), we use tensor parallelism to split the two embeddings across GPUs in the first pipeline stage, avoiding replicating position embeddings in Megatron-LM.
For (2), the next token prediction yields an intermediate activation of shape $[s, b, V]$, where $V$ is the vocabulary size, and is around 50k for a typical GPT family model.
To avoid stashing this activation, the next token prediction and loss computation is moved to the backward execution.
\section{Evaluation} \label{eval}
\subsection{Experiment Setup}
\textbf{Testbed.}  
The experiments were conducted on two GPU clusters with distinct hardware configurations.
The first cluster features H20 GPU nodes, each equipped with eight H20 GPUs and four InfiniBand NDR host channel adapters (HCAs) providing 200 Gbps bandwidth per port.
The second cluster consists of A800 GPU nodes.
They have eight A800 GPUs per node and are interconnected via four InfiniBand HDR HCAs, each operating at 100 Gbps.

\begin{table}[t!]
    \centering
    \caption{Targeting model configurations.}
    \begin{tabular}{cccc}
    \toprule
    Model Size & \#Layers & \#Heads & Hidden size \\ \midrule
    1.3B & 24 & 16 & 2048 \\
    3B & 16 & 32 & 4096 \\
    7B & 32 & 32 & 4096 \\ \bottomrule
    \end{tabular}
    \label{tab:model_params}
\end{table}

\begin{figure*}[!t]
    \centering
    \subfloat[Normalized throughput for 1.3B model on H20.]{
        \includegraphics[width=0.48\textwidth]{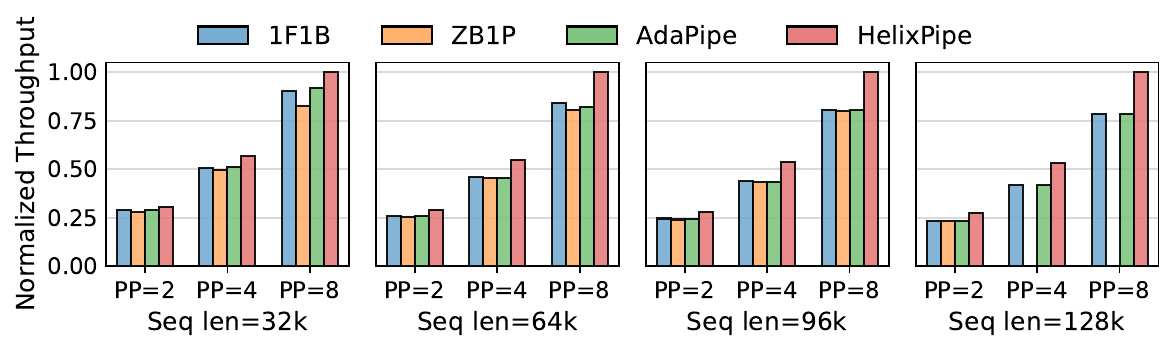}
    } \hfill
    \subfloat[Normalized throughput for 1.3B model on A800.]{
        \includegraphics[width=0.48\textwidth]{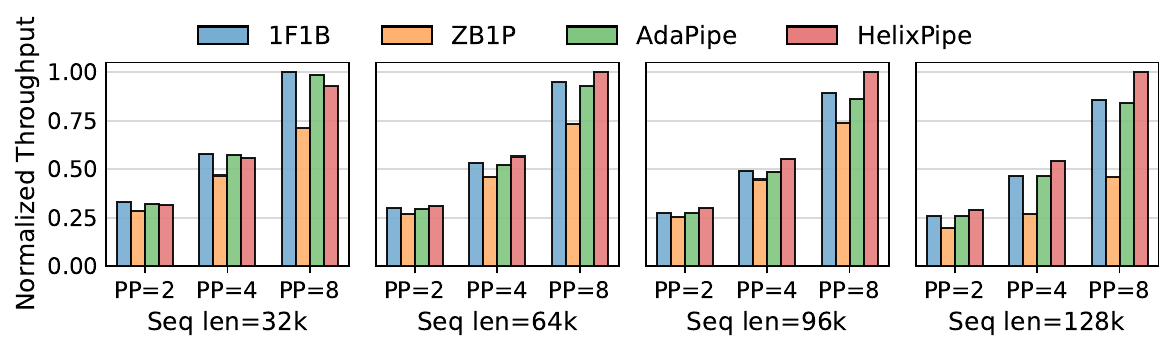}
    } \\
    \subfloat[Normalized throughput for 3B model on H20.]{
        \includegraphics[width=0.48\textwidth]{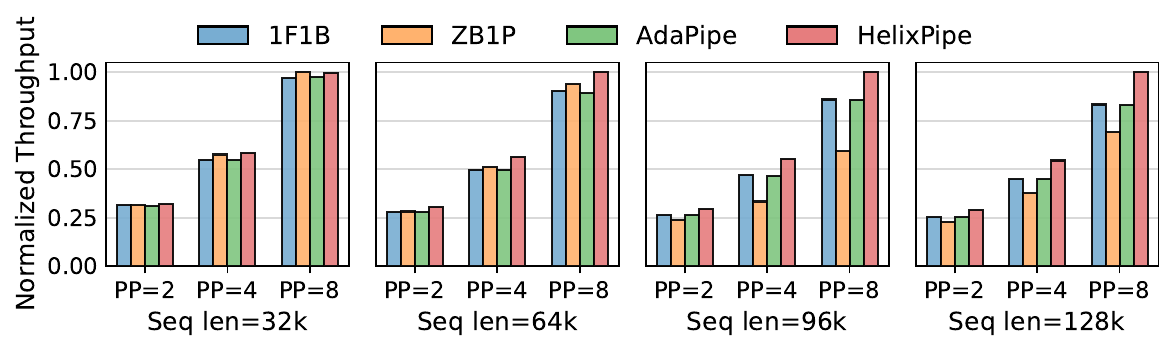}
    } \hfill
    \subfloat[Normalized throughput for 3B model on A800.]{
        \includegraphics[width=0.48\textwidth]{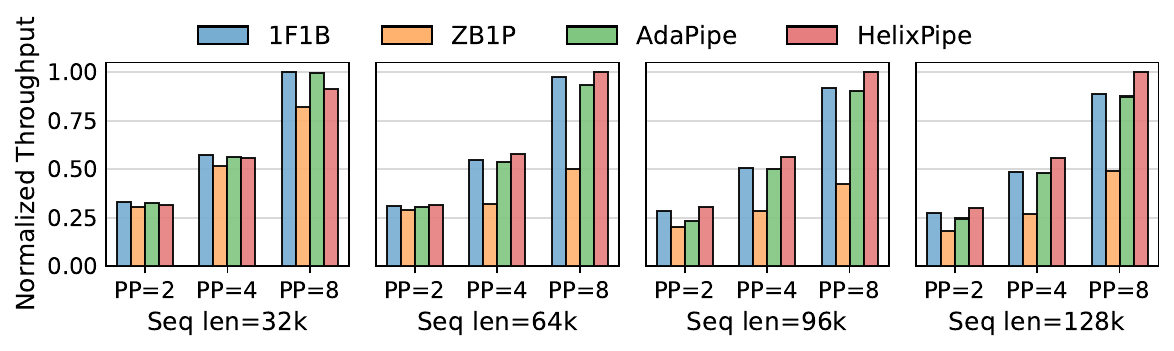}
    } \\
    \subfloat[Normalized throughput for 7B model on H20.]{
        \includegraphics[width=0.48\textwidth]{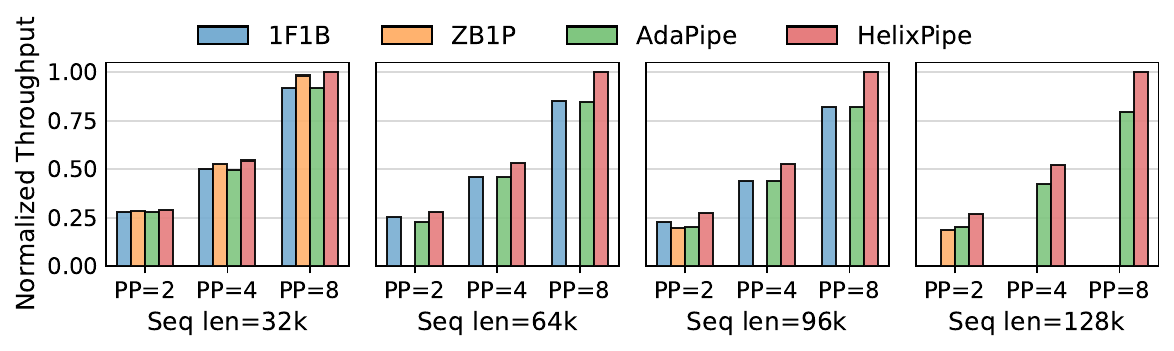}
    } \hfill
    \subfloat[Normalized throughput for 7B model on A800.]{
        \includegraphics[width=0.48\textwidth]{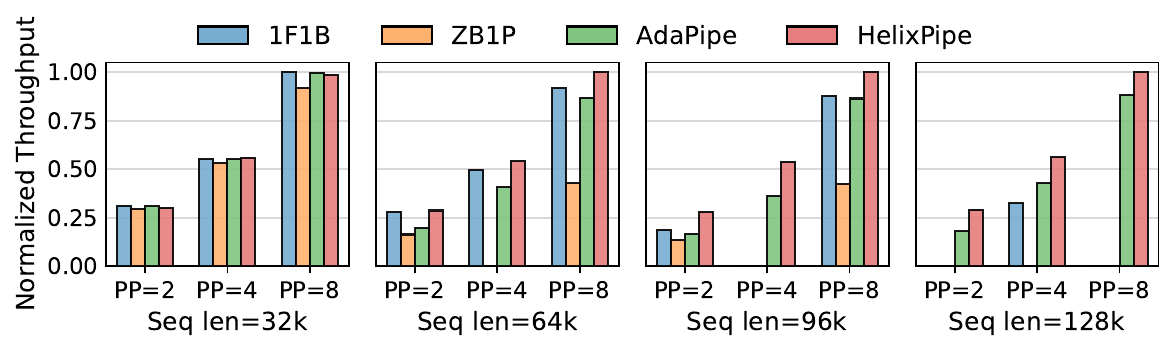}
    }
    \caption{Normalized throughput for different pipeline parallelisms, model scales, pipeline (cluster) sizes, and GPU types.}
    \label{fig:exp-throughput}
\end{figure*}

\textbf{Workloads.}
The targeting model architecture is the standard GPT-3 \cite{brown2020language}, and we varied the number of layers and the hidden size to test the performance with models of 1.3B, 3B and 7B parameters.
The detailed model configurations are listed in Table \ref{tab:model_params}.
For each configuration, we tested various long sequence lengths including 32k, 64k, 96k and 128k.
The experiments adopted synthesized datasets where each batch had the full targeting sequence lengths to rule out the effect of padding.

We conducted experiments at 2, 4 and 8 nodes to demonstrate the scalability of different pipeline parallelisms.
For each pipeline size, one pipeline stage was mapped to one node so the communication for pipeline parallelisms happened across nodes.
Within each node, we adopted Megatron sequence parallelism \cite{korthikanti2023reducing} and fixed the sequence parallel size to 8 to partition individual model parameters or activations and effectively exploit the fast NVLink for collective communications.
It should be noted that sequence parallelism and pipeline parallelism are orthorgonal to each other, so HelixPipe can also be integrated with other sequence parallelisms such as Deepspeed Ulysses \cite{jacobs2023deepspeed}, Ring attention \cite{liu2023ring, li2023sequence}, Megatron context parallelism \cite{context_parallel}, USP \cite{fang2024unified} to extend the sequence length at the intra-layer level.
The micro batch size was set to 1 and the global batch size was set to double the pipeline size to saturate the pipeline.
In addition, we enabled expandable segments by setting \texttt{PYTORCH\_CUDA\_ALLOC\_CONF} to mitigate the memory fragmentation issue with long sequence lengths for all methods \cite{guo2024gmlake}.

\textbf{Baselines.} 
The performance of HelixPipe was compared with three existing pipeline parallelisms, \textbf{1F1B} \cite{fan2021dapple, narayanan2021memory} and the zero bubble pipeline parallelism \textbf{ZB1P} \cite{qi2024zero}, which are discussed in Section \ref{pp-background}.
1F1B is the most adopted pipeline parallelism for training transformers \cite{llama3, bi2024deepseek}, and ZB1P has demonstrated its effectiveness to train transformers at scale \cite{deepseekv2, liu2024deepseek}.
To compare HelixPipe with memory-efficient pipeline parallelism, we reported results of \textbf{AdaPipe} \cite{sun2024adapipe}, which employs adaptive recomputation within each pipeline stage to maximize the memory utilization and adaptive partition to balance the computation across pipeline stages.
We applied the same optimizations mentioned in Section \ref{impl-opt} for all baseline methods to ensure a fair comparison.

Beyond the three baseline methods, we note that there are many other pipeline parallelisms, such as interleaved 1F1B \cite{narayanan2021efficient}, Chimera \cite{li2021chimera}, WeiPipe \cite{lin2025weipipe}, etc.
We extend their discussion and the reason we do not use them for experiments in Section \ref{related-work}.

\textbf{Metrics.}
We reported the normalized training throughput per iteration to measure the computation efficiency, which was obtained by averaging the execution time of 10 iterations after running 20 iterations to warm up.
We reported the maximum allocated memory by Pytorch's caching allocator to measure the peak memory overhead.

\subsection{Computation Efficiency and Scalability}
This section compares the training throughput to demonstrate the computation efficiency and the scalability on pipeline (cluster) size, sequence length, model scale and GPU type of different methods.
The experiment results are visualized in Figure \ref{fig:exp-throughput}.

Generally, compared to the best baseline method in each configuration, HelixPipe improves the training throughput by \textbf{28\%, 20\% and 26\%} when training the 1.3B, 3B and 7B models with 128k sequence length and 8 pipeline stages (nodes) on the H20 cluster.
On A800 cluster, HelixPipe speeds up the training by \textbf{16\%, 13\% and 13\%} for the corresponding settings.
For smaller cluster sizes and shorter sequence lengths, HelixPipe also shows comparable performance or descent speedup.

Specifically, Figure \ref{fig:exp-throughput} demonstrates the scalability of HelixPipe from three aspects.
The first scalability aspect is the \textbf{sequence length}.
With longer sequence length, the attention computation accounts for more fraction in the pipeline bubble time of baseline methods.
As a result, HelixPipe can show increasing performance benefits.
The second aspect of the scalability is the \textbf{model scale}.
For experiment results on the three model scales, the performance of HelixPipe maintains similar trends.
The last aspect is the \textbf{pipeline size}, or weak scalability.
As the pipeline size increases, HelixPipe shows consistent performance gain compared to baseline methods for all settings other than those with 32k sequence length and A800 GPU devices.
Also, it is noteworthy that the three scalability aspects of HelixPipe do not change on A800 cluster, but its speedup is smaller than it on H20 cluster.
The reasons are twofold.
First, A800 GPU has double computation power compared to H20 GPU, so the attention compuation is faster on A800 and the performance benefits from the A800 GPU also become smaller.
Second, A800 cluster only has half communication bandwidth than H20 cluster, which makes the communication bottleneck introduced by the two-fold FILO schedule cannot be overlapped by computation.
The impact of communication bandwidth on the two-fold FILO is further investigated in the following Section \ref{comm impact}.


\begin{figure}[!t]
    \centering
    \includegraphics[width=\columnwidth]{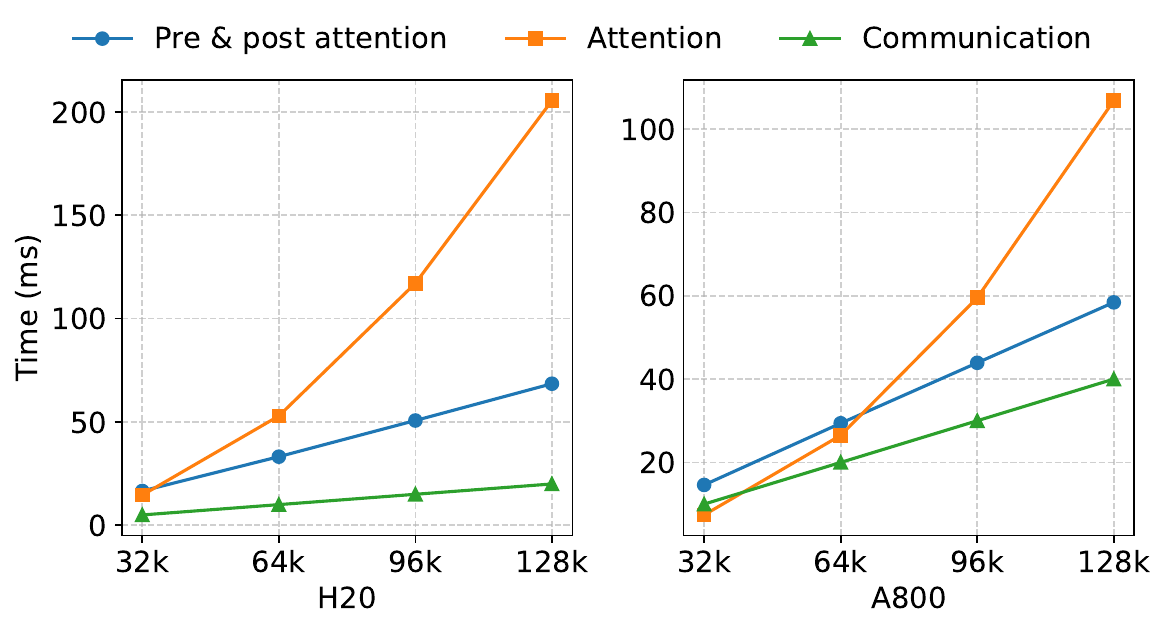}
    \caption{Decoupled computation time for a layer of the 7B model and the communication time costed by the two-fold FILO schedule with different sequence lengths.}
    \label{fig:exp-trace}
\end{figure}

For the three baseline methods, 1F1B shows the most stable performance, and it achieves the best results on A800 cluster with 32k sequence length.
AdaPipe is capable of handling longer sequence lengths and larger model scales with its adaptive recomputation strategy.
Nevertheless, its computation efficiency is no better than 1F1B in all cases. This is because, for long sequence lengths where attention dominates the computation each layer, there is almost no room to balance the pipeline stage computation via adaptive partition. 
Though theoretically, ZB1P should incur the same peak memory as 1F1B and should be no worse than 1F1B, it shows the most unstable computation results.
This is caused by the uneven computation between the backward \textit{B} and \textit{W} computation, which makes its scheduling difficult to find suitable position to place the backward \textit{W} computation.
Moreover, we observed severe memory fragmentation issues during its execution.

\begin{figure}[!t]
    \centering
    \includegraphics[width=0.9\columnwidth]{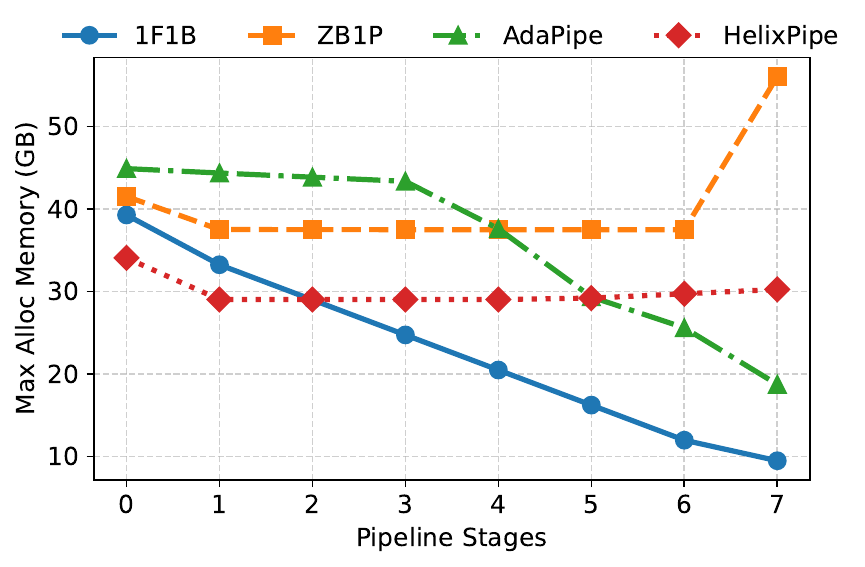}
    \caption{Memory overhead for different pipeline parallelisms training the 3B model with 128k sequence length on 8 pipeline stages.}
    \label{fig:memory}
\end{figure}

\subsection{Communication Impact on the Two-fold FILO Schedule} \label{comm impact}
To investigate the performance drop for HelixPipe with short sequence lengths on the A800 cluster and get more insights into the asynchronous two-fold FILO schedule, we present the decoupled computation and the estimated communication time in Figure \ref{fig:exp-trace}.
Specifically, this figure reports the forward execution time for the combined pre-attention and post-attention, and the attention for a layer of the 7B model. 
It also shows the estimated time for each p2p communication operation between pipeline stages under the two clusters. 
The overall communication volume is two activations as analyzed in Section \ref{attn-parallel-partition}.

This figure indicates that \textit{whether the two-fold FILO schedule works depends on if the attention computation can overlap the communication behind for the two-fold FILO schedule}.
Concretely, for A800 cluster, the attention computation for the sequence length of 32k is faster than the inter-node communication.
As a result, the communication introduced by the two-fold FILO schedule cannot be fully overlapped by the attention computation, leading to the delay of following computation.
Nevertheless, HelixPipe can still achieve descent speedup for longer sequence lengths on the A800 cluster, because the communication would be fully overlapped by the computation.
On the other hand, the communication on H20 cluster is fast, so it can be perfectly overlapped by the attention computation, which means that HelixPipe can scale to any number of nodes on this cluster.

It is noteworthy that the p2p communication is implemented with NCCL \cite{nccl}, which relies on GPU SMs to perform communication.
Such implementation may simultaneously delay the asynchronous communication and the computation.
However, we observed that there was only a marginal delay in computation time, and the communication was overlapped as expectation.
This is because only a small number of SMs is enough to fully exploit the communication bandwidth \cite{liu2024deepseek, zhang2024mpmoe}.

\subsection{Memory Footprint} \label{memory}
To demonstrate the memory overhead for different pipeline parallelisms, the memory footprint across eight pipeline stages for training the 3B model with 128k sequence length is shown in Figure \ref{fig:memory}.
Note that for the same configuration, the memory overhead is almost the same for the two clusters.
Besides, similar observations can be obtained for different number of pipeline stages.

The experiment results reveal that HelixPipe costs the lowest peak memory usage, and it shows the most balanced memory footprint across the eight pipeline stages.
1F1B consumes a skewed amount of memory for different pipeline stages and ZB1P maintains the same peak memory usage for all stages, as we have analyzed in Section \ref{motivation:compute}.
However, ZB1P incurs extremely high memory usage at the final stage.
This is due to the need for stashing the gradients of the word embeddings for next token prediction and loss computation.
Such memory is often stashed in fp32 format and ZB1P needs to save multiple micro batches for their backward \textit{W}.
AdaPipe achieves the goal of maximizing the memory utilization for first three stages, but fails for the following stages because the long sequence length makes it difficult to achieve the balance goal for adaptive partition.

\subsection{Recomputation Impact} \label{recomp impact}
\begin{figure}[!t]
    \centering
    \includegraphics[width=\columnwidth]{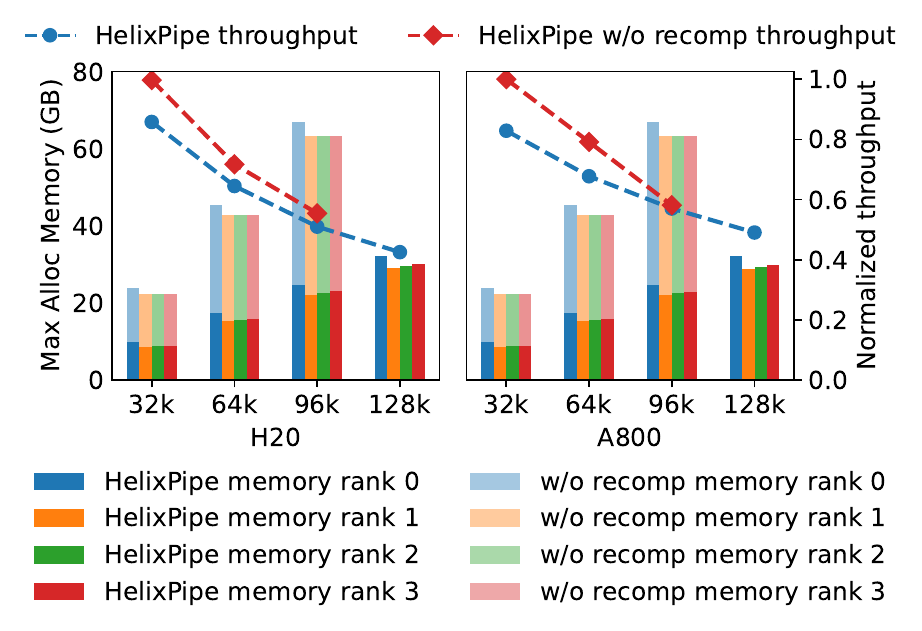}
    \caption{The memory footprint and normalized throughput with and without the recomputation without attention strategy training the 3B model with 4 pipeline stages.}
    \label{fig:recomp}
\end{figure}
The main benefits of HelixPipe on the memory side is from the recomputation without attention strategy.
To reveal its impact on both the computation and memory overhead, we show the experiment results with and without this strategy for training a 3B model with 4 pipeline stages on the two clusters in Figure \ref{fig:recomp}.

The results show that the recomputation without attention strategy would sacrifice the computation efficiency up to 20\% for shorter sequence lengths such as 32k or 64k sequence length.
However, as the sequence length increases, the throughput gap decreases and even gets near zero for 96k sequence length on the A800 cluster.
This is because the execution time for the combined pre-attention and post-attention accounts for a large portion of each layer when compared with the attention for short sequences, but the portion would decrease drastically with longer sequence lengths.
Moreover, after removing the recomputation without attention strategy, HelixPipe is unable to train longer sequence lengths beyond 128k.

\section{Related Work} \label{related-work}
\subsection{Long Sequence Transformer Training}
Flash attention optimizes the attention computation on a single GPU by reducing the IO between GPU HBM and on-chip memory \cite{dao2022flashattention}.
By lowering memory requirements from quadratic to linear scaling, it unlocks the potential for much longer sequence lengths.

Most existing work extends the sequence length with sequence parallelisms which split individual activations to multiple GPUs \cite{korthikanti2023reducing, jacobs2023deepspeed, fang2024unified}.
To partition an activation of shape $[s, b, h]$, there are two choices for sequence parallelisms.
The first one is to partition along sequence ($s$) dimension.
Ring self-attention \cite{li2023sequence, liu2023ring, li2023lightseq, liu2024wallfacerharnessingmultidimensionalring}, Megatron context parallelism \cite{context_parallel} and TeraPipe \cite{li2021terapipe} fall into this class.
They partition the activations of queries, keys and values along sequence dimension to different GPUs, and use p2p communication to transfer required key and value partitions to compute the attention output.
Another class of sequence parallelisms partition activations along the hidden size ($h$) dimension.
Methods including Megatron sequence parallelism \cite{korthikanti2023reducing}, Deepspeed Ulysses \cite{jacobs2023deepspeed} compute partial attention output and rely on collective communications to synchronize the final output.

Different from the above methods on the intra-layer level, WeiPipe resolves the long sequence transformer training from the inter-layer level with pipeline parallelism \cite{lin2025weipipe}, which is similar to HelixPipe.
However, it mainly focuses on reducing \textit{communication volume} via transferring model weights and their gradients instead of activations.
WeiPipe does not address the computation or memory overhead as HelixPipe targets on.
Besides, the two-fold FILO schedule validates that the communication introduced by pipeline parallelisms can be effectively overlapped.

\subsection{Computation-efficient Pipeline Parallelism}
The computation dependency between layers and the optimization step in each training iteration leads to pipeline bubble when using pipeline parallelisms.
The 1F1B pipeline in Megatron-LM is initially introduced by PipeDream \cite{narayanan2021memory} and DAPPLE \cite{fan2021dapple}.
It partitions a model into a consecutive layers for each stage, and executes micro batches following one-forward-one-backward schedule.
Instead of assigning a single set of consecutive layers to each stage, interleaved 1F1B reduces the pipeline bubble by partitioning the model to multiple chunks of layers, and each stage possesses a subset of chunks \cite{narayanan2021efficient}.
With more chunks, the pipeline bubble can be further reduced.
However, it requires extensive micro batches to saturate the pipeline when attaining to its theoretical minimum pipeline bubble rate, making it less desirable for long sequence training.
Chimera optimizes the pipeline bubble by duplicating the model states for each stage, and runs a bidirectional schedule to execute micro batches in two ends of stages \cite{li2021chimera}.
Hanayo eliminates the need of model state duplication with a wave-like schedule to simulate the execution of duplicated model states \cite{liu2023hanayo}.
Zero bubble pipeline decouples the gradient computation for input and parameters in the backward pass, and fills the bubble with the parameter gradient computation \cite{qi2024zero}.

GPipe also follows the FILO schedule to execute micro batches, but it simply partitions a model by layers to each stage \cite{huang2019gpipe}.
To reduce the pipeline bubble of GPipe, breadth-first pipeline introduces the interleaved model partition like interleaved 1F1B, but it follows the FILO schedule to achieve high GPU utilization with small batch size. \cite{lamy2023breadth}.
Out-of-order backprop adopts the idea of executing non critical operations out of order in single GPU, data parallelism and pipeline parallelism to improve the GPU utilization \cite{oh2022out}.
At the pipeline level, its core idea is equivalent to zero bubble pipeline but follows the FILO schedule.

As discussed in Section \ref{motivation:compute}, the partition strategy of the methods mentioned above is at the layer granularity, which makes the pipeline bubble proportional to the execution time of a layer.
With longer sequence length, the bubble time also increases.
HelixPipe breaks the boundary of layers with attention parallel partition to further reduce pipeline bubble and improve computation efficiency.

\subsection{Memory-efficient Pipeline Parallelism} 
While activation memory imbalance in 1F1B pipelines has been mentioned by various studies \cite{narayanan2021efficient, li2021chimera, liu2023hanayo}, few works have addressed this issue.
BPipe designs a communication schedule to balance activations across pipeline stages by transferring activations from early to later stages \cite{kim2023bpipe}.
However, this introduces additional data movement without computation efficiency benefits and can detrimentally affect communication efficiency when enabling data or tensor parallelism.
AdaPipe exploits adaptive activation recomputation to balance both the memory overhead and execution time at every pipeline stage of 1F1B \cite{sun2024adapipe}.
However, this work is mainly designed with up to 16k sequence length.
As discussed in Section \ref{motivation:compute}, when the sequence length is extremely long, it will dominate the computation time, leaving marginal space for AdaPipe to meet the balance.

HelixPipe adopts the FILO schedule to balance the memory footprint.
Furthermore, it utilizes recomputation without attention along with chunked MLP to relieve the memory pressure.

\section{Conclusion}
We present HelixPipe, a novel pipeline parallelism method for efficient transformer training with long sequences. 
By introducing attention parallel partition, HelixPipe removes attention computation from pipeline bubbles, improving computation efficiency. 
The two-fold FILO micro batch schedule balances memory usage across stages while overlapping communication with computation. 
Further optimizations, such as recomputation without attention and chunked MLP, reduce memory overhead. 
Experiments demonstrate that HelixPipe outperforms existing methods, achieving up to 26\% higher throughput when training a 7B model with 128k sequence length on 64 GPUs. 
Our work provides an effective solution for scaling transformer training to longer sequences.


\begin{acks}
We would like to acknowledge that computational work involved in this research work is supported by NUS IT’s Research Computing group using grant numbers NUSREC-HPC-00001.
\end{acks}

\bibliographystyle{ACM-Reference-Format}
\bibliography{ref}




\end{document}